\documentclass[hidelinks]{ecai} 

\usepackage{latexsym}
\usepackage{amssymb}
\usepackage{amsmath}
\usepackage{amsthm}
\usepackage{booktabs}
\usepackage{enumitem}
\usepackage{graphicx}
\usepackage{color}
\usepackage{algorithm}
\usepackage{algpseudocode}
\usepackage{orcidlink}
\usepackage{caption}
\usepackage{afterpage}
\usepackage[numbers]{natbib}

\DeclareMathOperator*{\argmin}{arg\,min}

\newcommand{\BibTeX}{B\kern-.05em{\sc i\kern-.025em b}\kern-.08em\TeX}

\begin{document}


\begin{frontmatter}


\paperid{2239} 


\title{Hyperparameter Importance Analysis for Multi-Objective AutoML}


\author[A,B]{\fnms{Daphne}~\snm{Theodorakopoulos}{\orcidlink{0009-0008-0827-7449}}\thanks{Corresponding Author. Email: d.theodorakopoulos@ai.uni-hannover.de}}
\author[A]{\fnms{Frederic}~\snm{Stahl}{\orcidlink{0000-0002-4860-0203}}}
\author[B,C]{\fnms{Marius}~\snm{Lindauer}{\orcidlink{0000-0002-9675-3175}}}

\address[A]{Marine Perception Research Department,
German Research Center for Artificial Intelligence (DFKI)}
\address[B]{Institute of Artificial Intelligence (LUH|AI), Leibniz University Hannover}
\address[c]{L3S Research Center}


\begin{abstract}
Hyperparameter optimization plays a pivotal role in enhancing the predictive performance and generalization capabilities of ML models. 
However, in many applications, we do not only care about predictive performance but also about additional objectives such as inference time, memory, or energy consumption.
In such multi-objective scenarios, determining the importance of hyperparameters poses a significant challenge due to the complex interplay between the conflicting objectives. 
In this paper, we propose the first method for assessing the importance of hyperparameters in multi-objective hyperparameter optimization. 
Our approach leverages surrogate-based hyperparameter importance measures, i.e., fANOVA and ablation paths, to provide insights into the impact of hyperparameters on the optimization objectives.
Specifically, we compute the a-priori scalarization of the objectives and determine the importance of the hyperparameters for different objective tradeoffs. 
Through extensive empirical evaluations on diverse benchmark datasets with three different objective pairs, each combined with accuracy, namely time, demographic parity loss, and energy consumption, we demonstrate the effectiveness and robustness of our proposed method. 
Our findings not only offer valuable guidance for hyperparameter tuning in multi-objective optimization tasks but also contribute to advancing the understanding of hyperparameter importance in complex optimization scenarios. 
\end{abstract}

\end{frontmatter}

\section{Introduction}
The selection of appropriate hyperparameter configurations significantly impacts a model's ability to capture underlying patterns in the data and produce accurate predictions.
Optimizing hyperparameters is one of the main focus areas of Automated Machine Learning (AutoML) \cite{hutter-book19a,bischl-dmkd23a}.
Given a use case and a model, it is usually unknown which hyperparameters are worth tuning to achieve a good performance. Providing insights into this (e.g., \cite{probst-jmlr19a,rijn-kdd18a,moussa-ml-24a}) is valuable since it allows the design of better configuration spaces and gives a better understanding of the learning dynamics of Machine Learning (ML) algorithms.
HyperParameter Importance (HPI) offers a systematic method to gain insights into the influence of hyperparameters on the model's performance.
Understanding and optimizing hyperparameters are crucial steps in building effective ML models.
They allow us to fine-tune our algorithms, improve performance, and achieve better generalization.
\begin{figure}[ht]
    \centering
    \includegraphics[width=\columnwidth]{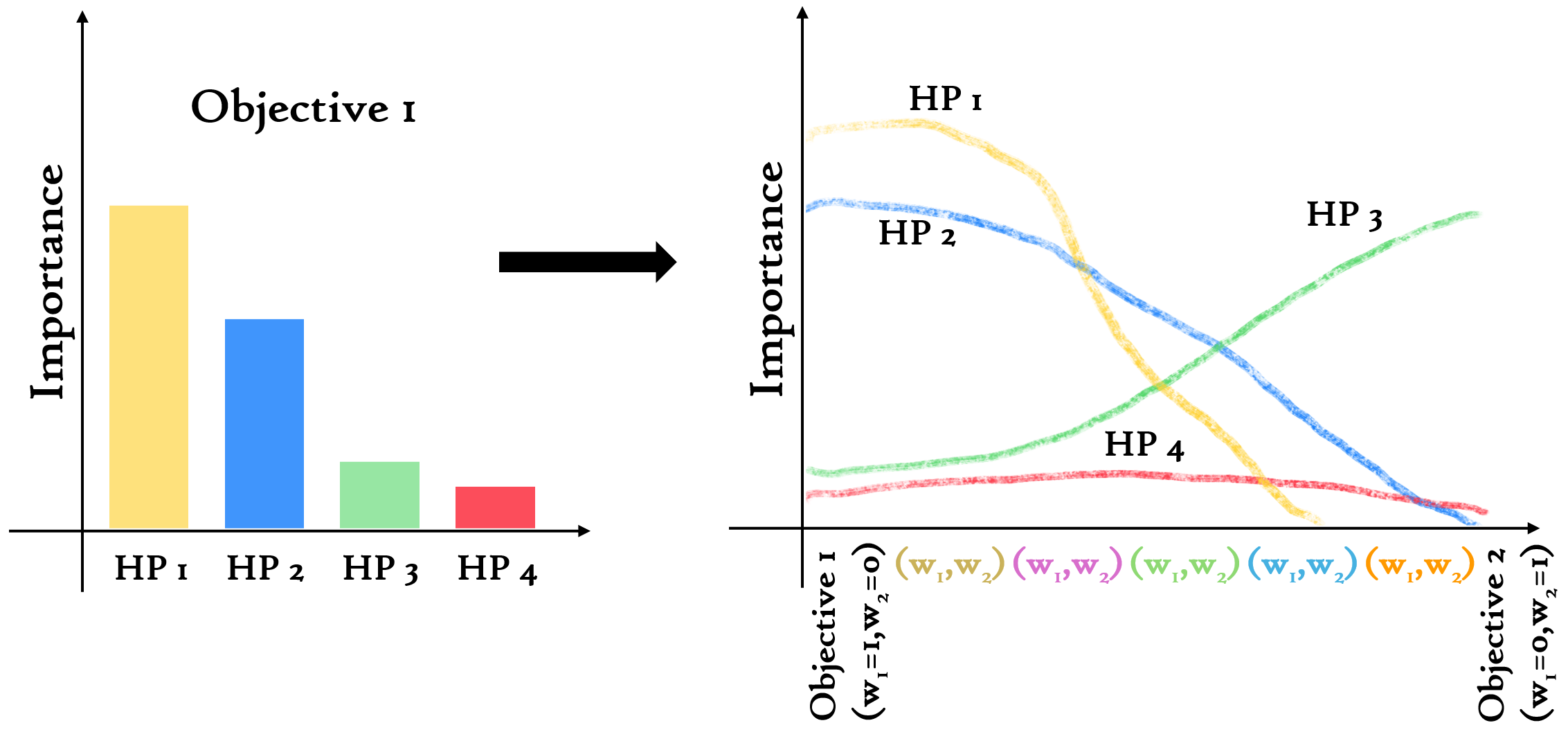}
    \caption{Overview of the MO-fANOVA method. On the left, the importance of each hyperparameter for Objective 1 is shown. On the right, our extension for the importance of each hyperparameter for different weightings of the objectives is displayed exemplarily.}
    \label{fig:fanova_overview}
\end{figure}
So far, most efforts in hyperparameter optimization (HPO) have been focused on single-objective optimization, mainly targeting the predictive performance of models. Recently, there has been a trend towards multi-objective HPO~\cite{hernandez-icml16a,belakaria-neurips19a,elsken-iclr19a,gardner-dsaa19a,binder-gecco20a,karl-evolearn23a}, which allows the consideration of several objectives, including fairness, memory consumption, training time, inference time, and energy consumption. In a Multi-Objective Optimization (MOO) scenario, where we optimize conflicting objectives simultaneously, it becomes more challenging to approximate the unknown Pareto front and to discern the relative significance of individual hyperparameters. The configurations on the Pareto front are the set of non-dominated solutions.

\begin{figure}[tbph]
    \centering
    \includegraphics[width=\columnwidth]{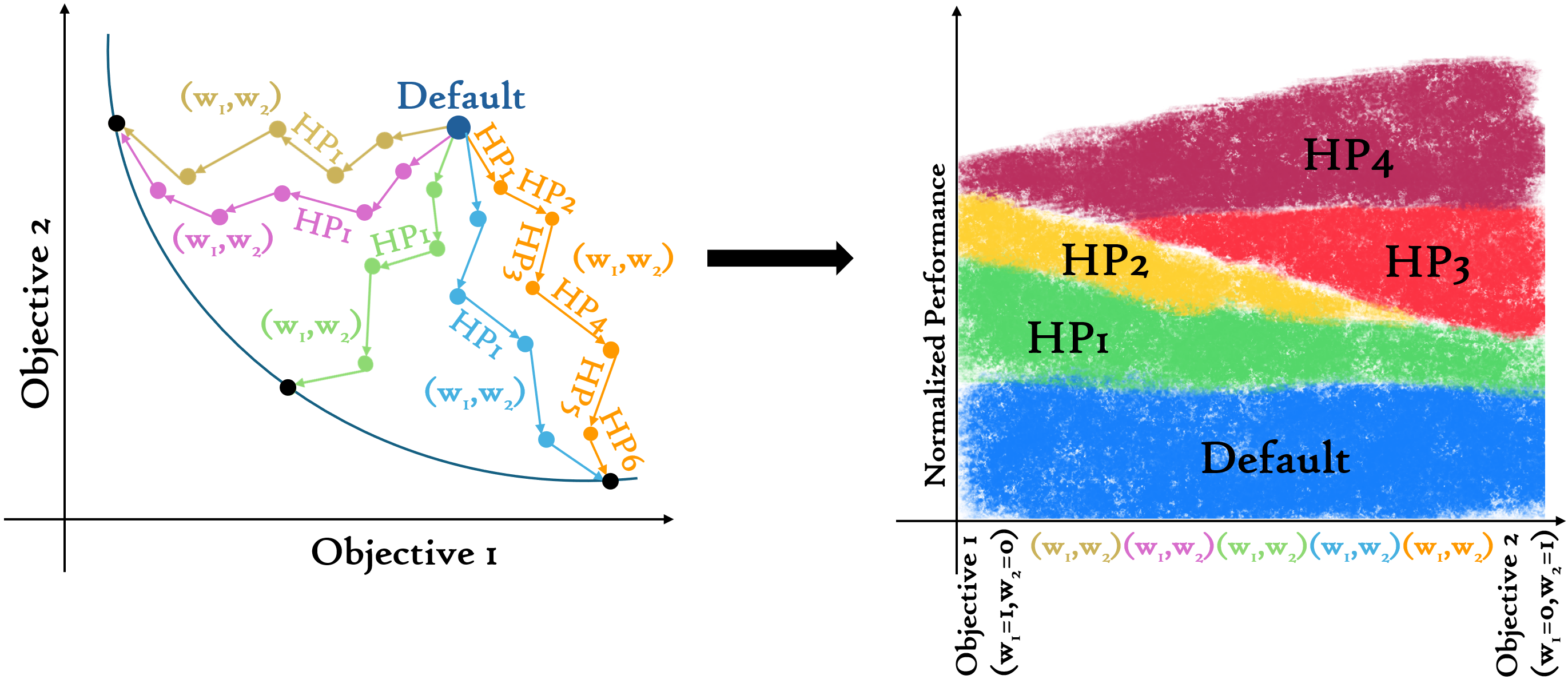}
    \caption{Overview of the MO-ablation path analysis. On the left, an exemplary Pareto front is displayed, with several ablation paths going from the default configuration to different configurations on the Pareto front. Every path is associated with a weighting of the objectives and thus gives a different value for the difference in performance per hyperparameter. We convert this to the plot on the right, where the total performance for different weightings is displayed as a stacked plot of hyperparameter contributions.}
    \label{fig:ablation_overview}
\end{figure}
Conventional methods for assessing HPI in single-objective scenarios rely on univariate sensitivity analysis techniques, like variance-based methods~\cite{hutter-icml14a} or partial dependence plots~\cite{moosbauer-neurips21a}. 
For instance, techniques like fANOVA have been widely used to decompose the variance of model performance into contributions from individual hyperparameters~\cite{hutter-icml14a}.
While effective for single-objective optimization, these methods may not capture the intricate interactions among multiple objectives in multi-objective settings.
So far, no methods have assessed the importance of hyperparameters within the context of MOO.

In this paper, we propose a novel method called weighted multi-objective HyperParameter Importance (MO-HPI). 
Although our approach is, in principle, usable with any surrogate-based HPI measure, we specifically focus on two widely-used methods, fANOVA~\cite{hutter-icml14a} and ablation path analysis \cite{fawcett-heu16a,biedenkapp-aaai17a}. 
Our methodology involves training surrogate models, e.g., random forests, with the hyperparameter configuration data and the respective objective results.
For better visualiziability and interpretability, we focus on bi-objective scenarios, but in principle, our approach can be used for any number of objectives.
We use weighted sums of the objectives as target variables, allowing us to effectively capture the change in importance from one objective to another.
Figure \ref{fig:fanova_overview} shows the concept of MO-fANOVA, and Figure \ref{fig:ablation_overview} the idea for the MO-ablation path analysis.

To analyze MO-HPI and demonstrate that it is reasonable in the context of ML, we pose the following research question:
\textbf{How to convert surrogate-based HPI methods to meaningfully measure the importance of hyperparameters within multi-objective optimization?} We further frame the following subquestions:
Are the results intuitively correct?
How do the proposed methods compare? 
Can we gain new insights into the impact of hyperparameters on the optimization process that previous methods could not provide?

To answer our research questions and validate the efficacy of our proposed method, we conduct three empirical evaluations on the well-known benchmark datasets MNIST \cite{deng-ieee-12a}, Adult Census Income \cite{becker-uci-96a} and CIFAR10 \cite{krizhevsky-tech09a}. 
To show that the approach can be applied to diverse objectives, we consider three pairs of objectives, each combined with accuracy: training time, demographic parity loss, and energy consumption, and discuss the results accordingly. 

By advancing the understanding of HPI in complex optimization scenarios, our method offers a valuable technique that can aid researchers and practitioners in tackling challenging MOO problems more effectively.

\section{Related Work}
Several techniques are available for evaluating the importance of hyperparameters. 
\textbf{Surrogate models}, such as Gaussian process models \cite{rasmussen-book06a} or random forests \cite{breiman-mlj01a}, have been employed to approximate the relationship between hyperparameters and performance.
They can predict the performance of a given hyperparameter configuration based on an empirical dataset of configurations and their performances.
\citet{breiman-mlj01a} introduced how random forests can attribute importance by observing performance changes when removing attributes. 
Forward selection \cite{hutter-lion13a} uses this idea by selecting the hyperparameters that most affect the surrogate's performance by starting from no hyperparameters and iteratively adding the most impactful ones. 
Local Parameter Importance \cite{biedenkapp-lion18a} studies the performance changes of a configuration along each hyperparameter by considering the variance in performance when changing the hyperparameter.
Recent works \cite{adachi-arxiv23a,rodemann-arxiv24a} use Shapley values \cite{shapley-book53a} to measure the HPI in Bayesian optimization inspired by the fact that Shapley values can quantify the hyperparameter attributions for the acquisition function.

Another method is \textbf{ablation path analysis} \cite{fawcett-heu16a,biedenkapp-aaai17a}, which compares the default and optimized configuration to measure hyperparameter contributions.
It creates an ablation path from the default configuration to a target configuration by changing the hyperparameter with the largest increase in performance in each iteration. We note that these ablation paths do not follow traditional ablation studies in which only a single hyperparameter is changed, and all others are fixed; however, this allows a full path through the configuration space from the default configuration to a target configuration.
While there is only one path in a single-objective scenario, potentially, there are many in the MOO setting (cf. Figure \ref{fig:ablation_overview}).

\textbf{fANOVA} \cite{hutter-icml14a} identifies the importance of individual hyperparameters and interactions among them. 
For each hyperparameter, it measures how it contributes to the variance in performance.
This is done, e.g., by training a random forest as a surrogate model and subsequently decomposing the variance of each tree into contributions to each subset of hyperparameters.
Based on these methods, several works have been published. 
For example, \citet{rijn-kdd18a}, \citet{probst-jmlr19a} and \citet{moussa-ml-24a} considered HPI across datasets.
PED-ANOVA \cite{watanabe-ijcai-23a} improved fANOVA using Pearson divergence to work better on arbitrary subspaces of the search space, e.g., the subspace of the top-performing configurations.
This paper contributes to this body of work by extending the methodology to MOO.

\section{Weighted HPI for Multiple Objectives}
Our main idea is simple and intriguing, given the insight that we are interested in the HPI for different objective tradeoffs. The main challenge lies in (i) how to map this to different HPI methods, (ii) how to compute this efficiently, (iii) how to visualize this, and (iv) how to interpret this. Challenges (i)-(iii) are described in this section, and (iv) follows as part of the following experiment and discussion sections.
Overall, our work employs a framework for assessing the influence of hyperparameters on multiple objectives using any surrogate-based HPI analysis methods, allowing it to be efficient in computing analyses. 
Given a performance meta-dataset obtained by a MO-HPO optimizer, we train surrogate models to predict the objectives on unevaluated configurations; this is then used to calculate the HPI.
The steps will be explained in more detail in the following section. While we discuss our approaches for two competing objectives, the approach can be straightforwardly extended to more objectives. The implementation is available on GitHub at \href{https://github.com/automl/hpi_for_mo_automl}{https://github.com/automl/hpi\_for\_mo\_automl}.

In the following, we use the following notation. $(O_1,O_2)$ denote two objective values. $(\tilde{O_1},\tilde{O_2})$ are the normalized versions of them. We use $\vec{W}$ to denote a vector of weights weighting the objectives. If we sum the weighted objectives, we obtain $Y_w$. All the evaluated configurations are collected in $\Lambda$, where $\lambda \in \Lambda$ denotes one evaluated configuration. $hps$ are hyperparameter names and $hp_{\text{min}}$ is the most important one. To get the predicted (weighted) objective values $r$ of any configuration, we train two surrogate models $\mathcal{S}_{obj}$, each trained with $\Lambda$ and $O_{\text{obj}}$ for the ablation path analysis, weighted after prediction. For fANOVA, we train one surrogate model per weighting $\mathcal{S}_w$ with $\Lambda$ and $Y_w$ (weighted before training).

\subsection{Data Preparation.}
The performance meta-dataset contains model configurations with their performance on several objectives.
That means a table where each row contains hyperparameter configurations with the respective performance of all measured objectives.
First, each objective is normalized using the min-max normalization across all evaluated configurations.
Normalization is done to scale objectives, such as time, so that all objectives live on the same scale and thus are comparable in their magnitude.
Next, the configuration data is converted to numerical values to deal with categorical data and NaN values.
NaN values could occur because of hierarchical structures based on conditional hyperparameters, i.e., hyperparameters that are only part of the configuration if a certain value of another hyperparameter is chosen.

\subsection{Weighting Scheme}
Each configuration on the Pareto front implicitly refers to one weighted tradeoff of the objectives.
Only Pareto-efficient objective pairs $(o_1,o_2)$ are considered to retrieve the corresponding weighting.
Subsequently, the normalized objective pairs are scaled to sum up to 1 by adding each pair up and dividing each of the two values by that sum to be used as weighting ($\tilde{o_1} = \frac{o_1}{o_1 +o_2}$).

\subsection{Multi-Objective fANOVA}
\begin{algorithm}[ht]
\caption{Multi-Objective fANOVA}
\hspace*{\algorithmicindent} \textbf{Input:} evaluated and encoded configurations $\lambda \in \Lambda$, corresponding evaluated objective values $(O_1,O_2)$ 
\begin{algorithmic}[1]
    \State $\text{$(\tilde{O_1},\tilde{O_2}) \gets$ normalized objective values}$
    \State $\text{$\vec{W} \gets$ is\_pareto\_efficient($(\tilde{O_1},\tilde{O_2})$)}$
    \For{$\text{each $\vec{w}$ in $\vec{W}$}$}
        \State $Y_w \gets w_1 \cdot \tilde{O_1} + w_2 \cdot \tilde{O_2}$
        \State $\text{$\mathcal{S}_w \gets$ surrogate model trained with $(\Lambda, Y_w)$}$
        \State $\text{Calculate fANOVA importances with $\mathcal{S}_w$, e.g. following \cite{hutter-icml14a}}$
    \EndFor
\end{algorithmic}
\label{alg:mo_fanova}
\end{algorithm}


A single target variable (Y) is created for each weighting $(w_1,w_2)$ derived from the evaluated configurations $\lambda \in \Lambda$ on the Pareto front by summing the weighted objectives values $(O_1,O_2)$: 
\begin{equation}
    Y_w = w_1 \cdot O_1 + w_2 \cdot O_2
    \label{eq:target}
\end{equation}

Given the weighting and the prepared data, MO-fANOVA can be calculated. 
Algorithm \ref{alg:mo_fanova} shows the procedure in detail.
A probabilistic surrogate model is trained for each weighting.
The model takes the encoded configurations as input and the weighted normalized sum of the objectives as the target variable.
After that, the fANOVA importance is calculated using the trained model for the corresponding weighting. See Figure~\ref{fig:fanova_overview} for an exemplary depiction of the outcome.

\subsection{Multi-Objective Ablation Path Analysis}
As shown by Biedenkapp et al.~\cite{biedenkapp-aaai17a}, ablation paths, i.e., the path of flipping the value of a hyperparameter from a given default configuration to a target configuration, can be efficiently approximated by using a surrogate model trained on meta-data collected by algorithm configuration or HPO. However, the standard procedure of computing these ablation paths by a greedy scheme~\cite{fawcett-heu16a} only considers a single objective.

\begin{algorithm}[ht]
\caption{Multi-Objective Ablation Path Analysis}
\hspace*{\algorithmicindent} \textbf{Input:} evaluated and encoded configurations $\lambda \in \Lambda$, corresponding evaluated objective values $(O_1,O_2)$, hyperparameters $hps$ 
\begin{algorithmic}[1]
\State $\text{$\vec{W} \gets$ is\_pareto\_efficient$((O_1,O_2))$}$
\For{$\text{each $obj$ in $(O_1,O_2)$}$} \Comment{Not normalized} \label{line:train_mdoels}
    \State $\text{$\mathcal{S}_{obj} \gets$ surrogate model trained with $\Lambda, O_{\text{obj}}$}$
\EndFor
\For{$\text{each $\vec{w}$ in $\vec{W}$}$} \label{line:ablation_path}
    \State $\lambda_{\text{best}} \gets \underset{\Lambda}{\argmin}(w_1 \cdot O_1 + w_2 \cdot O_2)$   \label{line:incumbent}
    \State $r_{\text{previous}} \gets w_1 \cdot \mathcal{\tilde{S}}_1(\lambda_{\text{default}}) + w_2 \cdot \mathcal{\tilde{S}}_2(\lambda_{\text{default}}))$
    \State $r_{\text{min}} \gets r_{\text{previous}}$
    \State $\lambda_{\text{previous}} \gets \lambda_{\text{default}}$
    \State $\lambda_{\text{current}} \gets \lambda_{\text{previous}}$
    \While{$r_{\text{min}} \leq r_{\text{previous}}$} \label{line:while} \Comment{As long as we find improvements}
        \For{$hp$ in $hps$}  \label{line:abl_start}
            \State $\lambda_{\text{current}}[hp] \gets \lambda_{\text{best}}[hp]$
            \State $r \gets w_1 \cdot \mathcal{\tilde{S}}_1(\lambda_{\text{current}}) + w_2 \cdot \mathcal{\tilde{S}}_2(\lambda_{\text{current}}))$
            \label{line:normalize1}
            \If{$r < r_{\text{min}}$} \label{line:max_diff1}
                \State $hp_{\text{min}} \gets hp$ \Comment{Most important hyperparameter}
                \State $r_{\text{min}} \gets r_{\text{total}}$
            \EndIf \label{line:max_diff2}
            \State $\lambda_{\text{current}} \gets \lambda_{\text{previous}}$
        \EndFor \label{line:abl_end}
        \State Add $hp_{\text{min}}$ with $r_{\text{min}}$ to ablation path 
        \State $\lambda_{\text{previous}}[hp_{\text{min}}] = \lambda_{\text{best}}[hp_{\text{min}}]$ \label{line:max_diff}
        \State Remove $hp_{\text{min}}$ from $hps$
    \EndWhile
\EndFor
\end{algorithmic}
\label{alg:mo_ablation}
\end{algorithm}

We propose to extend the ablation path analysis for multiple objectives as follows; see also Algorithm \ref{alg:mo_ablation}. 
For both objectives, a surrogate model is trained with the encoded configuration data and the objectives before normalization (Line \ref{line:train_mdoels}).
For each weighting, the ablation path is calculated (starting from line \ref{line:ablation_path}).
The incumbent $\lambda_\text{best}$ is the corresponding configuration to the minimum of the scalarized target variable $Y_w$ of Equation \ref{eq:target} (Line \ref{line:incumbent}).
Starting from the default configuration, each hyperparameter value is changed individually to the value of the incumbent configuration, and the performance is estimated with the trained surrogate models (Lines \ref{line:abl_start}-\ref{line:abl_end}).
The results of the models are normalized with the same normalization as the respective objective values, weighted, and summed up as performance (Line \ref{line:normalize1}).
The hyperparameter value leading to the highest difference in performance (toward the optimization objective) will be changed in the configuration (Lines \ref{line:max_diff1}-\ref{line:max_diff2} and line \ref{line:max_diff}).
The ablation is repeated with the altered configuration $\lambda_\text{current}$ and the remaining hyperparameters until the incumbent configuration is reached (Line~\ref{line:while}).
The difference in performance is recorded for each weighting.

\section{Experiments}
Using our multi-objective HyperParameter Importance (MO-HPI) approach, we evaluated three HPO problems with different data, models, and objectives.
The overview can be seen in Table \ref{tab:experiments}.
We chose rather simple benchmarks and models so that the results of the MO-HPI approach are easier to validate by common knowledge.

\begin{table}[ht]
\caption{Overview of the experiments}
\resizebox{\columnwidth}{!}{%
\centering
\begin{tabular}{l|rrrr}
\toprule
\textbf{Name} & \textbf{Dataset} & \textbf{Model} & \textbf{Objective 1} & \textbf{Objective 2} \\ \midrule
\emph{time} & MNIST & MLP & 1-Accuracy & Training Time \\ 
\emph{fairness} & Adult Census & MLP & 1-Accuracy & Demographic Parity Loss \\ 
\emph{energy} & CIFAR10 & ResNet & 1-Accuracy & Energy Consumption \\ \bottomrule
\end{tabular}
}
\label{tab:experiments}
\end{table}

\begin{figure*}[ht]
    \centering
    \includegraphics[width=\textwidth]{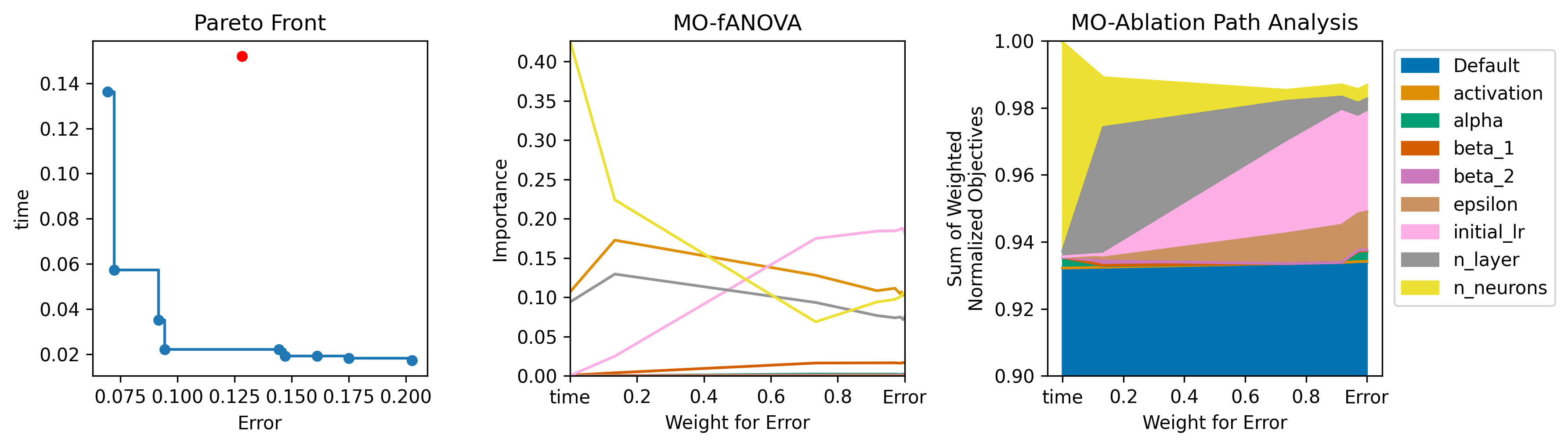}
    \caption{Results for the \emph{time} experiment. The Pareto front is on the left (error vs. training time in seconds), with the red dot being the default performance. The MO-fANOVA results are in the middle, and the MO-ablation path analysis is on the right. The x-axis corresponds to the weighting of the minimum error objective.}
    \label{fig:time}
\end{figure*}

\subsection{General Setup}
The basic setup was the same for all experiments. 
We used SMAC \cite{hutter-lion11a,lindauer-jmlr22a} as one of the state-of-the-art HPO tools~\cite{eggensperger-neuripsdbt21a} to perform HPO.
The multi-objective algorithm ParEGO \cite{knowls-evoco06a} with Random Online Aggressive Racing (ROAR) \cite{hutter-lion11a} and 1000 trials was used for optimization.
ROAR randomly selects a configuration from the hyperparameter space and only keeps track of the top 10 non-dominated configurations. We note that our primary goal was not to achieve the best possible multi-objective performance but to use a sufficient amount of performance data generated by an HPO tool. Since Moosbauer et al.~\cite{moosbauer-neurips21a} showed that post hoc analysis of HPO runs can be biased by too strong exploitation, we focus here on unbiased randomly generated data. 
For both HPI methods, we trained random forests with 100 trees as surrogate models.
Note that reproducing our experiments will lead to slightly different results for the \emph{time} and \emph{energy} experiment, as those variables depend on factors such as the hardware.

\subsection{Experiment: Time} \label{sec:time}

\begin{table}[ht]
\caption{Hyperparameter Space for Multi-Layer Perceptron}
\resizebox{\columnwidth}{!}{%
    \centering
    \begin{tabular}{l|rrrr}
        \toprule
        \textbf{Hyperparameter} & \textbf{Range} & \textbf{Scale} & \textbf{Default} \\
        \midrule
        n\_layer & 1-5 & linear &  3 \\
        
        \begin{tabular}[c]{@{}l@{}} n\_neurons (same\\ for all layers)\end{tabular} & 8-256 & log &  132 \\
        
        activation & logistic, tanh, relu & - &  tanh \\
        
        initial\_lr & 0.0001-0.1 & log&  0.01 \\
        
        alpha & 0.0001-1.0 & log&  0.1 \\
        
        beta\_1 & 0.1-1.0 & log&  0.5 \\
        
        beta\_2 & 0.1-1.0 & log&  0.5 \\
        
        epsilon & 1e-10-1e-06 & log&  1e-8 \\
        \bottomrule
    \end{tabular}
    }
\label{tab:hps_mlp}
\end{table}

The first experiment used the MNIST dataset with Multi-layer Perceptron (MLP) classifiers based on Sci-kit learn's implementation~\cite{scikit-learn}.
The objective ``training time'' was measured by the wallclock time the classifier used for training.
The error was measured by 1 minus the accuracy on the test set.
The MLPs were configured with 50 maximum epochs and Adam as the DL optimizer.
SMAC tuned the MLPs hyperparameter w.r.t. the configuration space in Table \ref{tab:hps_mlp}.
Figure \ref{fig:time} shows the Pareto front (left), the MO-fANOVA (middle), and the MO-ablation path analysis (right) results for the \emph{time} experiment.
The Pareto front shows the set of non-dominated solutions. The red point represents the default configuration.
Note that for the x-axis, only the weighting for the error objective is displayed. Since the total weighting always sums to 1, the opposite weighting applies to the respective other objective — in this case, training time, so $w_2$ = $1-w_1$.
MO-fANOVA thus displays the HPI from a low weighting of the error, so a high weighting of the time objective, to a high weighting of the error and a low weighting of time.
For example, the number of neurons starts at an importance of around 0.4 when only considering the time objective and decreases the more the error objective is weighed in.
The ablation path analysis plot is stacked, with each segment representing the performance increase attributed to tuning a specific hyperparameter for each weighting. 
The blue part shows the performance of the default configuration.
Note that the y-axis only starts at 0.9, which means the default configuration already does quite well without tuning the hyperparameters.

\subsection{Experiment: Fairness}

\begin{figure*}
    \centering
    \includegraphics[width=\textwidth]{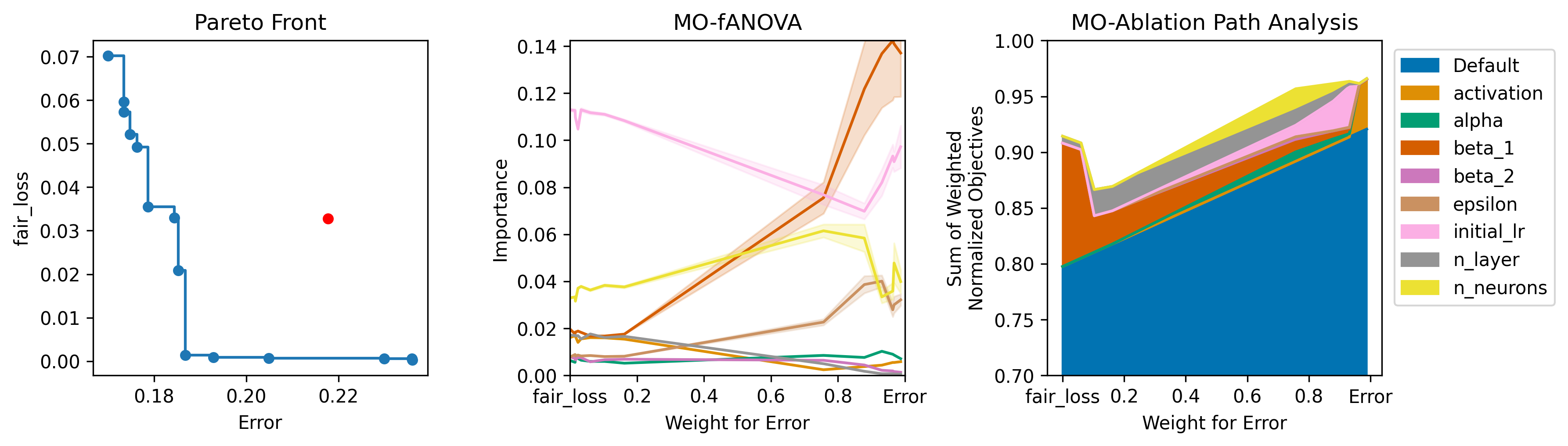}
    \caption{Results for the \emph{fairness} experiment. The Pareto front is on the left (error vs. demographic parity loss), with the red dot being the default performance. The MO-fANOVA results are in the middle, and the MO-ablation path analysis is on the right. The x-axis corresponds to the weighting of the minimum error objective.}
    \label{fig:fair_loss}
\end{figure*}

The second experiment used the Adults Census Income dataset \cite{becker-uci-96a}, which contains several variables about US citizens with the binary target variable annual income higher or lower than 50'000\$.
We only used the numeric and binary nominal variables of the dataset.
It contains the sensitive variables ``sex'' and ``race''.
We calculated the fairness loss as the second objective for the experiment based on the ``race'' variable prediction using demographic parity (DP).
It is calculated by the absolute difference of the mean proportions of positive predictions $y$ in each group, where the groups are defined by a sensitive variable $s$.

\begin{equation}
    DP Loss = \left| \frac{\sum_{i=1}^{n} y_{_i}(s_i = 0)}{n}   - \frac{\sum_{i=1}^{n} y_{_i}(s_i = 1)}{n} \right|
\end{equation}

The MLPs were set up the same as in the \emph{time} experiment.
Figure \ref{fig:fair_loss} shows the Pareto front, the MO-fANOVA, and the MO-ablation path analysis results for the \emph{fairness} experiment.
While the optimization found several fair configurations, an error remained, with a top accuracy of around 83\%.
The plots can be interpreted the same way as described in section \ref{sec:time}.

\subsection{Experiment: Energy Consumption}

\afterpage{%
    \begin{figure*}[ht!]
        \centering
        \includegraphics[width=\textwidth]{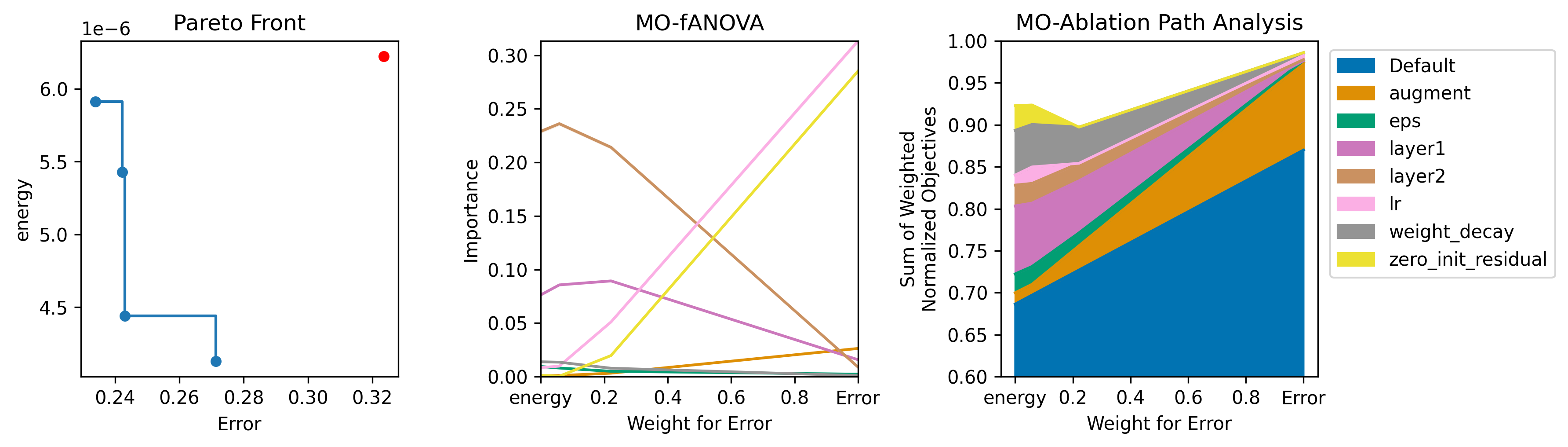}
        \caption{Results for the \emph{energy} experiment. The Pareto front is on the left (error vs. inference energy consumption), with the red dot being the default performance. The MO-fANOVA results are in the middle and the MO-ablation path analysis is on the right. The x-axis corresponds to the weighting of the minimum error objective.}
        \label{fig:energy}
    \end{figure*}
}

Our last experiment used the Torchvision implementation of ResNets \cite{he-cvpr16a} with the first and the last layer set to a size of 3.
The models were trained with AdamW and a cosine annealing learning rate scheduler with 200 maximum iterations.
They were trained for 50 epochs with early stopping, which monitored the test accuracy with a patience of 6 epochs.
The CIFAR10 dataset was always normalized with mean and standard deviation.
Optionally, the data was augmented with a random crop of size 32 and a random horizontal flip.
The objective ``inference energy'', so for predicting the test set, was estimated with CodeCarbon~\citep{DBLP:journals/corr/abs-1910-09700,DBLP:journals/corr/abs-1911-08354}\footnote{https://github.com/mlco2/codecarbon}.
The configuration space is displayed in Table~\ref{tab:hps_resnet}.
Figure \ref{fig:energy} shows the Pareto front, the MO-fANOVA, and the MO-ablation path analysis results for the \emph{energy} experiment.
The plots can be interpreted the same way as described in section \ref{sec:time}.

\begin{table}
\caption{Hyperparameter Space for ResNet}
\resizebox{\columnwidth}{!}{%
    \centering
    \begin{tabular}{l|rrrr}
        \toprule
        \textbf{Hyperparameter} & \textbf{Belongs to} & \textbf{Range} & \textbf{Scale} & \textbf{Default} \\
        \midrule
        layer1 & Model & 1-30 & linear & 15 \\ 
        layer2 & Model & 1-30 & linear & 15 \\ 
        zero\_init\_residual & Model & true or false & - & true \\ 
        augment & Dataset & true or false & - & false \\ 
        learning rate & Optimizer & 0.0001 - 0.1 & log & 0.01 \\ 
        weight\_decay & Optimizer & 0.00001 - 0.1 & log & 0.001 \\ 
        eps & Optimizer & 1e-10-1e-06 & log & 1e-8 \\ \bottomrule
        \end{tabular}
}
\label{tab:hps_resnet}
\end{table}

\section{Discussion}
Our research aimed to \textbf{investigate whether surrogate-based HyperParameter Importance (HPI) methods could be converted to meaningfully measure HPI within MOO}.
We introduced the concept of weighting the objectives based on configurations along the Pareto front and applied it to two HPI methods: fANOVA and ablation path analysis.
Subsequently, we conducted three MO-HPO experiments with diverse datasets, models, and objectives to assess the usefulness of the two MO-HPI approaches.
In the following, we discuss our research questions related to the experimental results from the last section.

\subsection{Are the results intuitively correct?}
The results show the effectiveness of our method.
All three experiments provide plots that make sense from the ML perspective, and thus, we conjecture that the results are intuitively correct.
This can be seen in Figure \ref{fig:time} for the ablation and \ref{fig:energy} for fANOVA, where hyperparameters related to network size (such as layer size and number of layers) strongly influence training time and energy consumption, consistent with the understanding that larger neural networks require more energy and time.
A similar observation can be made for the learning rate.
Its impact on the error is to be expected, given its well-known importance for achieving high accuracy. 
Additionally, in the second experiment, the learning rate plays a role in fairness (cf. Figure \ref{fig:fair_loss}). 
The same is valid for data augmentation in the \emph{energy} experiment, as is shown in Figure \ref{fig:energy}, it has a high importance, and it is known to heavily influence performance \cite{wang-arxiv17a}.
However, some unexpected results, like the high importance of beta\_1 in the \emph{fairness} experiment, are not intuitively explainable and warrant further exploration. We note that recent results showed that improved fairness can be strongly related to hyperparameters~\cite{dooley-fairnas-2023}. 
Overall, we therefore claim that the results are intuitively correct, which answers our first research question positively.

\subsection{How do the proposed methods compare?}
When comparing the two methods, i.e., MO-fANOVA and MO-ablation path analysis, some interesting observations can be made.
In the \emph{time} experiment, the activation function is less important when measured by the ablation path analysis than by fANOVA, but epsilon gains more importance for the error objective in the ablation path analysis.
Another surprising observation is that in the \emph{fairness} experiment, beta\_1 is deemed important by both measures, but for the fANOVA measure, it has a strong influence on the error side, and for the ablation path analysis, it has a stronger influence on the fairness.
Another difference is the high importance of the learning rate in Figure \ref{fig:fair_loss} for fANOVA. At the same time, there is not much contribution of the learning rate visible for the ablation path analysis.
In the \emph{energy} experiment, in the ablation path analysis, data augmentation rated high, and the learning rate has minimal relevance in comparison to fANOVA.

The difference in the results of the methods is mostly due to their distinct nature.
It is well known that the ablation paths struggle with correctly attributing importance in case of interactions of hyperparameters~\cite{fawcett-heu16a}. Since the value of two or more hyperparameters has to be changed before an effect on the performance can be observed, the importance is only attributed to the last changed hyperparameter.
Consequently, the performance differences upon changing the other hyperparameters might be lower than they actually are. 
In the single objective scenario, this is partially mitigated by also displaying the order of changes, which is not possible with our visualization.

Furthermore, the ablation results strongly depend on the chosen default configuration as a reference point. If the default configuration is already fairly close to one of the extreme ends of the Pareto front, there are little performance gains that can be attributed to the hyperparameters. Some hyperparameters might even seem unimportant because they are already set to the optimal value concerning the dominating objective. Therefore, it is important to always interpret the ablation path in view of the default configuration.
Since fANOVA for main effects studies the variance for each hyperparameter independent of the others, it is more stable in that sense. Our method also allows us to study the importance of higher-order interaction effects with fANOVA, as in the single-objective case. To avoid cluttering our plots and results, we have omitted this here. 

Lastly, it is important to note that ablation paths are considered a local HPI analysis since their path through the configuration space potentially only covers a small part of it and can be seen as a local interpretation of the incumbent configuration given a default configuration. In contrast, fANOVA covers the entire configuration space and thus provides more high-level insights, which, on the other hand, might not be important for specific performance improvements w.r.t. the default configuration.
Overall, it is helpful to use both methods, as they provide diverse insights.

\subsection{Can we gain new insights that previous methods could not provide?}
We could simply run HPI studies on each objective independently and compare their results. We argue that by considering different tradeoffs of the objectives -- as typically done in MOO -- we gain valuable insights from our methods that were not possible previously. 

Considering the tradeoffs between the objectives, we can make the following observations.
In the fANOVA plot of the \emph{time} experiment (cf. Figure \ref{fig:time}), most hyperparameters significantly impacting only one of the objectives intersect at approximately an equal weighting of the objectives.
This suggests that the MO-fANOVA in this case contains as much information as calculating fANOVA independently. Nevertheless, even in Figure~\ref{fig:time}, it is not a trivial linear trend of the importance values of the hyperparameters over different objective tradeoffs. 
In Figure \ref{fig:fair_loss} for the \emph{fairness} experiment, the intersections are still at around the same position but are not in the center of the plot. 
For the \emph{energy} experiment (cf. Figure \ref{fig:energy}), the intersections are even at different x-positions (i.e., objective weightings).

In addition, it is evident that certain hyperparameters lose importance rapidly (e.g., zero\_init\_residual in Figure \ref{fig:energy} for the ablation path analysis), while others remain important for longer before they become irrelevant (e.g., beta\_1 in Figure \ref{fig:fair_loss} for the ablation path analysis). 
Moreover, some hyperparameters would not be tuned when only looking at the objectives independently. 
For instance, in the ablation path analysis in Figure \ref{fig:fair_loss} (right), the initial learning rate only contributes significantly to the performance for certain tradeoffs but not to one of the extrema of the Pareto front. 

Furthermore, it is interesting to observe that certain hyperparameters consistently retain importance, for instance, in Figure \ref{fig:fair_loss}, for fANOVA, both the initial learning rate and the number of neurons remain influential. 
Conversely, some hyperparameters remain unimportant across all weightings.
This consistency provides clarity regarding whether these hyperparameters require tuning.
Finally, the stacked visualization makes it possible to easily see the relative contribution of a hyperparameter for different points on the Pareto front, which make up the weighting.
Therefore, we conclude that there is more information in the plots that consider different objective tradeoffs compared to analyzing the objectives independently.

\subsection{Limitations and Future Work}
As is always the problem with post-hoc analysis, although our results are reasonable, the absence of ground truth data prevents us from guaranteeing their correctness. Nevertheless, we believe that there is sufficient empirical evidence allowing us to conjecture that our method is correct. Although several approaches for studying (single-objective) results of HPO are already presented~\cite{hutter-icml14a,biedenkapp-aaai17a,moosbauer-neurips21a,watanabe-ijcai-23a,adachi-arxiv23a,rodemann-arxiv24a}, a systematic set of criteria as proposed for interpretable ML~\cite{nauta-cs23a} is still missing in the context of interpretable AutoML.

In our experiments, we have not explored MOO scenarios with more than two objectives because visualization and interpretability would become complex. Nonetheless, our approach can be extended to more objectives. Consider that the weighting scheme is calculated by dividing each coordinate value of a point on an n-dimensional Pareto front with the total sum of all values of that point, with each objective previously normalized between 0 and 1. For both methods, the target Y can now be calculated as the weighted sum of objectives with which the surrogate models can be trained. Plotting the results for three objectives in a 3D plot would be possible, but not trivially beyond that. We note that many-objective optimization (i.e., more than three objectives) could be relevant in practice but lead to a large fraction of the configurations being on the Pareto front. To the best of our knowledge, there are no reasonable approaches for many-objective HPO to date.

In future work, MO-AutoML could incorporate the proposed method to actively choose the most important hyperparameters to tune. This could be done in the MO setting by calculating the integrals of the MO-HPI over the different tradeoffs. This provides a way to quantify the importance one-dimensionally without having to decide on a tradeoff. This will lead to more efficient AutoML that learns to actively consider the HPI for different tasks, datasets, and models.
Moreover, it is already known that post-hoc analysis with partial dependence plots of HPO runs can be skewed because HPO approaches, such as Bayesian optimization~\cite{shahriari-ieee16a}, tradeoff exploitation and exploration and thus bias the sampled configurations accordingly~\cite{moosbauer-neurips21a}. So far, it is not known how this affects approaches such as fANOVA and ablation paths, but it is reasonable to assume that they are similarly negatively affected. Therefore, future work has to include how this will affect our approach and new approaches for how to fix skewed results under biased data.
In addition, our methods should be integrated into interactive tools, such as DeepCave \cite{sass-realml22a} or IOHprofiler~\cite{doerr-arxiv18b}. 
Another functionality could be enabling users to select configurations on the Pareto front and convert them into objective weightings.

\section{Conclusion}
In multi-objective hyperparameter optimization, assessing the importance of hyperparameters is challenging due to conflicting objectives. 
Our proposed method leverages surrogate-based hyperparameter importance measures, specifically fANOVA and ablation paths, in conjunction with a-priori scalarization across various objective tradeoffs.
While we focus on those two HPI methods, our approach can also work with other HPI methods, such as Partial Dependence Plots~\cite{moosbauer-neurips21a}, symbolic regression~\cite{segel-automl23a}, variants of fANOVA~\cite{watanabe-ijcai-23a}, or maybe even with visualization methods such as parallel coordinate plots~\cite{golovin-kdd17a} or configuration footprints~\cite{biedenkapp-lion18a,rasulo-gecco24a}.
Empirical evaluations demonstrate the effectiveness and interpretability of our method.
We believe our proposed approach will enhance the analysis of multi-objective AutoML results and thus contribute to a human-centered AutoML paradigm~\cite{lindauer-icml24a}.


\begin{ack}
The DFKI Niedersachsen (DFKI NI) is funded in the "zukunft.niedersachsen" by the Lower Saxony Ministry of Science and Culture and the Volkswagen Foundation (funding no. ZN3480). Marius Lindauer acknowledges support by the German Federal Ministry of the Environment, Nature Conservation, Nuclear Safety and Consumer Protection (GreenAutoML4FAS project no. 67KI32007A) and was funded by the European Union (ERC, ``ixAutoML'', grant no.101041029). Views and opinions expressed are however those of the author(s) only and do not necessarily reflect those of the European Union or the European Research Council Executive Agency. Neither the European Union nor the granting authority can be held responsible for them.
\includegraphics[height=5cm]{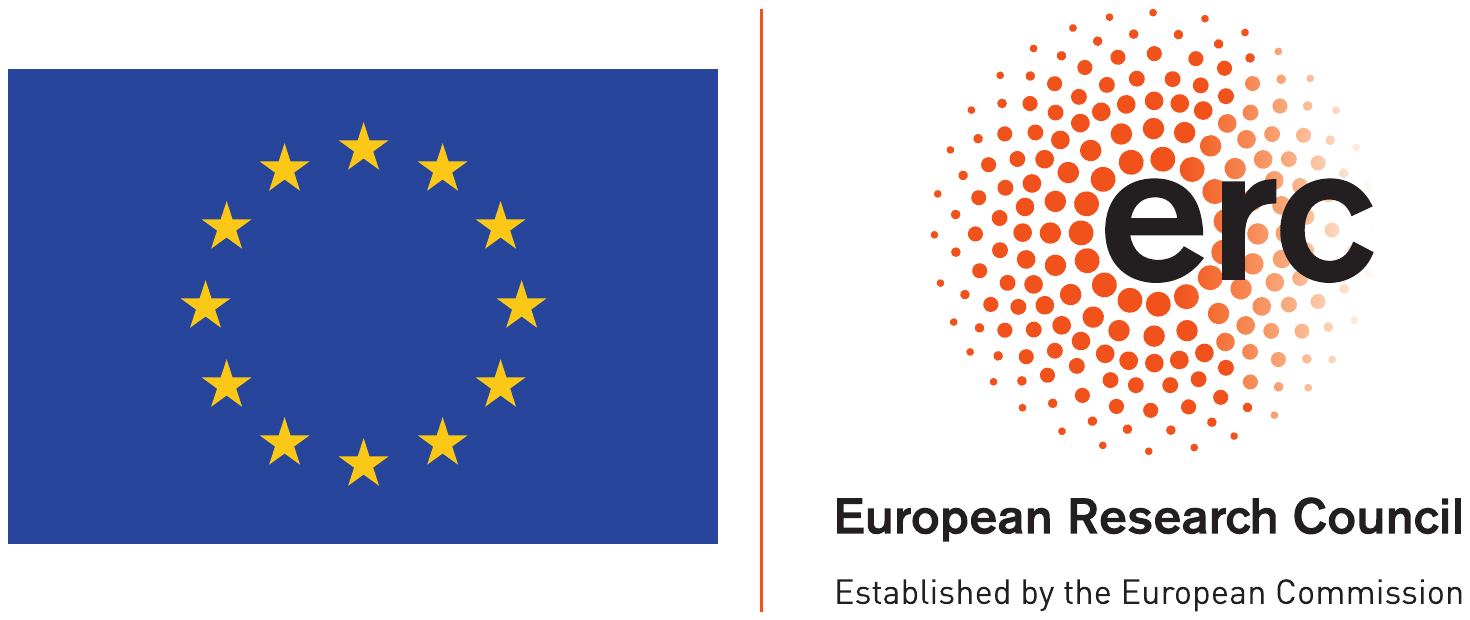}
\end{ack}



\bibliography{shortstrings,lib,mybibfile,proc}

\begin{thebibliography}{44}
\providecommand{\natexlab}[1]{#1}
\providecommand{\url}[1]{\texttt{#1}}
\expandafter\ifx\csname urlstyle\endcsname\relax
  \providecommand{\doi}[1]{doi: #1}\else
  \providecommand{\doi}{doi: \begingroup \urlstyle{rm}\Url}\fi

\bibitem[Adachi et~al.(2023)Adachi, Planden, Howey, Muandet, Osborne, and Chau]{adachi-arxiv23a}
M.~Adachi, B.~Planden, D.~Howey, K.~Muandet, M.~Osborne, and S.~Chau.
\newblock Looping in the human: Collaborative and explainable bayesian optimization.
\newblock \emph{CoRR}, abs/2310.17273, 2023.

\bibitem[Becker and Kohavi(1996)]{becker-uci-96a}
B.~Becker and R.~Kohavi.
\newblock {Adult}.
\newblock UCI Machine Learning Repository, 1996.
\newblock {DOI}: https://doi.org/10.24432/C5XW20.

\bibitem[Belakaria et~al.(2019)Belakaria, Deshwal, and Doppa]{belakaria-neurips19a}
S.~Belakaria, A.~Deshwal, and J.~Doppa.
\newblock Max-value entropy search for multi-objective {Bayesian} optimization.
\newblock In H.~Wallach, H.~Larochelle, A.~Beygelzimer, F.~d'Alche Buc, E.~Fox, and R.~Garnett, editors, \emph{Proceedings of the 33rd International Conference on Advances in Neural Information Processing Systems ({N}eur{IPS}'19)}. Curran Associates, 2019.

\bibitem[Biedenkapp et~al.(2017)Biedenkapp, Lindauer, Eggensperger, Fawcett, Hoos, and Hutter]{biedenkapp-aaai17a}
A.~Biedenkapp, M.~Lindauer, K.~Eggensperger, C.~Fawcett, H.~Hoos, and F.~Hutter.
\newblock Efficient parameter importance analysis via ablation with surrogates.
\newblock In S.~Singh and S.~Markovitch, editors, \emph{Proceedings of the Thirty-First Conference on Artificial Intelligence ({AAAI}'17)}, pages 773--779. {AAAI} Press, 2017.

\bibitem[Biedenkapp et~al.(2018)Biedenkapp, Marben, Lindauer, and Hutter]{biedenkapp-lion18a}
A.~Biedenkapp, J.~Marben, M.~Lindauer, and F.~Hutter.
\newblock {CAVE}: Configuration assessment, visualization and evaluation.
\newblock In R.~Battiti, M.~Brunato, I.~Kotsireas, and P.~Pardalos, editors, \emph{Proceedings of the International Conference on Learning and Intelligent Optimization ({LION})}, Lecture Notes in Computer Science. Springer-Verlag, 2018.

\bibitem[Binder et~al.(2020)Binder, Moosbauer, Thomas, and Bischl]{binder-gecco20a}
M.~Binder, J.~Moosbauer, J.~Thomas, and B.~Bischl.
\newblock Multi-objective hyperparameter tuning and feature selection using filter ensembles.
\newblock In J.~Ceberio, editor, \emph{Proceedings of the Genetic and Evolutionary Computation Conference ({GECCO}'20)}, pages 471--–479. ACM Press, 2020.

\bibitem[Bischl et~al.(2023)Bischl, Binder, Lang, Pielok, Richter, Coors, Thomas, Ullmann, Becker, Boulesteix, Deng, and Lindauer]{bischl-dmkd23a}
B.~Bischl, M.~Binder, M.~Lang, T.~Pielok, J.~Richter, S.~Coors, J.~Thomas, T.~Ullmann, M.~Becker, A.~Boulesteix, D.~Deng, and M.~Lindauer.
\newblock Hyperparameter optimization: Foundations, algorithms, best practices, and open challenges.
\newblock \emph{Wiley IRDMKD}, page e1484, 2023.

\bibitem[Breiman(2001)]{breiman-mlj01a}
L.~Breiman.
\newblock Random forests.
\newblock \emph{MLJ}, 45:\penalty0 5--32, 2001.

\bibitem[Deng(2012)]{deng-ieee-12a}
L.~Deng.
\newblock The mnist database of handwritten digit images for machine learning research.
\newblock \emph{IEEE Signal Processing Magazine}, 29\penalty0 (6):\penalty0 141--142, 2012.

\bibitem[Doerr et~al.(2018)Doerr, Wang, Ye, van Rijn, and B{\"a}ck]{doerr-arxiv18b}
C.~Doerr, H.~Wang, F.~Ye, S.~van Rijn, and T.~B{\"a}ck.
\newblock {IOH}profiler: A benchmarking and profiling tool for iterative optimization heuristics.
\newblock \emph{arXiv:1810.05281 [cs.NE]}, 2018.

\bibitem[Dooley et~al.(2023)Dooley, Sukthanker, Dickerson, White, Hutter, and Goldblum]{dooley-fairnas-2023}
S.~Dooley, R.~S. Sukthanker, J.~P. Dickerson, C.~White, F.~Hutter, and M.~Goldblum.
\newblock Rethinking bias mitigation: Fairer architectures make for fairer face recognition.
\newblock In \emph{Thirty-seventh Conference on Neural Information Processing Systems (NeurIPS 2023)}, 2023.

\bibitem[Eggensperger et~al.(2021)Eggensperger, M{\"u}ller, Mallik, Feurer, Sass, Klein, Awad, Lindauer, and Hutter]{eggensperger-neuripsdbt21a}
K.~Eggensperger, P.~M{\"u}ller, N.~Mallik, M.~Feurer, R.~Sass, A.~Klein, N.~Awad, M.~Lindauer, and F.~Hutter.
\newblock {HPOBench}: A collection of reproducible multi-fidelity benchmark problems for {HPO}.
\newblock In J.~Vanschoren and S.~Yeung, editors, \emph{Proceedings of the Neural Information Processing Systems Track on Datasets and Benchmarks}. Curran Associates, 2021.

\bibitem[Elsken et~al.(2019)Elsken, Metzen, and Hutter]{elsken-iclr19a}
T.~Elsken, J.~Metzen, and F.~Hutter.
\newblock Efficient multi-objective {N}eural {A}rchitecture {S}earch via lamarckian evolution.
\newblock In \emph{Proceedings of the International Conference on Learning Representations ({ICLR}'19)}, 2019.
\newblock Published online: \url{iclr.cc}.

\bibitem[Fawcett and Hoos(2016)]{fawcett-heu16a}
C.~Fawcett and H.~Hoos.
\newblock Analysing differences between algorithm configurations through ablation.
\newblock \emph{Journal of Heuristics}, 22\penalty0 (4):\penalty0 431--458, 2016.

\bibitem[Gardner et~al.(2019)Gardner, Golovidov, Griffin, Koch, Thompson, Wujek, and Xu]{gardner-dsaa19a}
S.~Gardner, O.~Golovidov, J.~Griffin, P.~Koch, W.~Thompson, B.~Wujek, and Y.~Xu.
\newblock Constrained multi-objective optimization for automated machine learning.
\newblock In L.~Singh, R.~{De Veaux}, G.~Karypis, F.~Bonchi, and J.~Hill, editors, \emph{Proceedings of the International Conference on Data Science and Advanced Analytics ({DSAA'19})}, pages 364--373. IEEE, {IEEE}, 2019.

\bibitem[Golovin et~al.(2017)Golovin, Solnik, Moitra, Kochanski, Karro, and Sculley]{golovin-kdd17a}
D.~Golovin, B.~Solnik, S.~Moitra, G.~Kochanski, J.~Karro, and D.~Sculley.
\newblock Google {V}izier: A service for black-box optimization.
\newblock In S.~Matwin, S.~Yu, and F.~Farooq, editors, \emph{Proceedings of the 23rd {ACM} {SIGKDD} International Conference on Knowledge Discovery and Data Mining ({KDD}'17)}, pages 1487--1495. ACM Press, 2017.

\bibitem[He et~al.(2016)He, Zhang, Ren, and Sun]{he-cvpr16a}
K.~He, X.~Zhang, S.~Ren, and J.~Sun.
\newblock Deep residual learning for image recognition.
\newblock In \emph{Proceedings of the International Conference on Computer Vision and Pattern Recognition ({CVPR}'16)}, pages 770--778. CVF and IEEE CS, IEEE, 2016.

\bibitem[Hern\'{a}ndez-Lobato et~al.(2016)Hern\'{a}ndez-Lobato, Hern\'{a}ndez-Lobato, Shah, and Adams]{hernandez-icml16a}
D.~Hern\'{a}ndez-Lobato, J.~Hern\'{a}ndez-Lobato, A.~Shah, and R.~Adams.
\newblock Predictive entropy search for multi-objective bayesian optimization.
\newblock In M.~Balcan and K.~Weinberger, editors, \emph{Proceedings of the 33rd International Conference on Machine Learning ({ICML}'17)}, volume~48, page 1492–1501. Proceedings of Machine Learning Research, 2016.

\bibitem[Hutter et~al.(2011)Hutter, Hoos, and Leyton-Brown]{hutter-lion11a}
F.~Hutter, H.~Hoos, and K.~Leyton-Brown.
\newblock Sequential model-based optimization for general algorithm configuration.
\newblock In C.~Coello, editor, \emph{Proceedings of the Fifth International Conference on Learning and Intelligent Optimization ({LION}'11)}, volume 6683 of \emph{Lecture Notes in Computer Science}, pages 507--523. Springer-Verlag, 2011.

\bibitem[Hutter et~al.(2013)Hutter, Hoos, and Leyton-Brown]{hutter-lion13a}
F.~Hutter, H.~Hoos, and K.~Leyton-Brown.
\newblock Identifying key algorithm parameters and instance features using forward selection.
\newblock In P.~Pardalos and G.~Nicosia, editors, \emph{Proceedings of the Seventh International Conference on Learning and Intelligent Optimization ({LION}'13)}, volume 7997 of \emph{Lecture Notes in Computer Science}, pages 364--381. Springer-Verlag, 2013.

\bibitem[Hutter et~al.(2014)Hutter, Hoos, and Leyton-Brown]{hutter-icml14a}
F.~Hutter, H.~Hoos, and K.~Leyton-Brown.
\newblock An efficient approach for assessing hyperparameter importance.
\newblock In E.~Xing and T.~Jebara, editors, \emph{Proceedings of the 31th International Conference on Machine Learning, ({ICML}'14)}, pages 754--762. Omnipress, 2014.

\bibitem[Hutter et~al.(2019)Hutter, Kotthoff, and Vanschoren]{hutter-book19a}
F.~Hutter, L.~Kotthoff, and J.~Vanschoren, editors.
\newblock \emph{Automated Machine Learning: Methods, Systems, Challenges}.
\newblock Springer, 2019.
\newblock Available for free at \url{http://automl.org/book}.

\bibitem[Karl et~al.(2023)Karl, Pielok, Moosbauer, Pfisterer, Coors, Binder, Schneider, Thomas, Richter, Lang, Garrido-Merchán, Branke, and Bischl]{karl-evolearn23a}
F.~Karl, T.~Pielok, J.~Moosbauer, F.~Pfisterer, S.~Coors, M.~Binder, L.~Schneider, J.~Thomas, J.~Richter, M.~Lang, E.~Garrido-Merchán, J.~Branke, and B.~Bischl.
\newblock Multi-objective hyperparameter optimization -- an overview.
\newblock \emph{Transactions of Evolutionary Learning and Optimization}, 3\penalty0 (4):\penalty0 1--–50, 2023.

\bibitem[Knowles(2006)]{knowls-evoco06a}
J.~D. Knowles.
\newblock {ParEGO}: a hybrid algorithm with on-line landscape approximation for expensive multiobjective optimization problems.
\newblock \emph{{IEEE} Transactions on Evolutionary Computation}, 10\penalty0 (1):\penalty0 50--66, 2006.

\bibitem[Krizhevsky(2009)]{krizhevsky-tech09a}
A.~Krizhevsky.
\newblock Learning multiple layers of features from tiny images.
\newblock Technical report, University of Toronto, 2009.

\bibitem[Lacoste et~al.(2019)Lacoste, Luccioni, Schmidt, and Dandres]{DBLP:journals/corr/abs-1910-09700}
A.~Lacoste, A.~Luccioni, V.~Schmidt, and T.~Dandres.
\newblock Quantifying the carbon emissions of machine learning.
\newblock \emph{arXiv:1910.09700}, 2019.

\bibitem[Lindauer et~al.(2022)Lindauer, Eggensperger, Feurer, Biedenkapp, Deng, Benjamins, Ruhkopf, Sass, and Hutter]{lindauer-jmlr22a}
M.~Lindauer, K.~Eggensperger, M.~Feurer, A.~Biedenkapp, D.~Deng, C.~Benjamins, T.~Ruhkopf, R.~Sass, and F.~Hutter.
\newblock {SMAC3}: A versatile bayesian optimization package for {H}yperparameter {O}ptimization.
\newblock \emph{JMLR}, 23\penalty0 (54):\penalty0 1--9, 2022.

\bibitem[Lindauer et~al.(2024)Lindauer, Karl, Klier, Moosbauer, Tornede, M{\"u}ller, Hutter, Feurer, and Bischl]{lindauer-icml24a}
M.~Lindauer, F.~Karl, A.~Klier, J.~Moosbauer, A.~Tornede, A.~M{\"u}ller, F.~Hutter, M.~Feurer, and B.~Bischl.
\newblock Position paper: A call to action for a human-centered automl paradigm.
\newblock In \emph{Proceedings of the international conference on machine learning}, 2024.

\bibitem[Lottick et~al.(2019)Lottick, Susai, Friedler, and Wilson]{DBLP:journals/corr/abs-1911-08354}
K.~Lottick, S.~Susai, S.~Friedler, and J.~Wilson.
\newblock Energy usage reports: Environmental awareness as part of algorithmic accountability.
\newblock \emph{arXiv:1911.08354}, 2019.

\bibitem[Moosbauer et~al.(2021)Moosbauer, Herbinger, Casalicchio, Lindauer, and Bischl]{moosbauer-neurips21a}
J.~Moosbauer, J.~Herbinger, G.~Casalicchio, M.~Lindauer, and B.~Bischl.
\newblock Explaining hyperparameter optimization via partial dependence plots.
\newblock In M.~Ranzato, A.~Beygelzimer, K.~Nguyen, P.~Liang, J.~Vaughan, and Y.~Dauphin, editors, \emph{Proceedings of the 35th International Conference on Advances in Neural Information Processing Systems ({N}eur{IPS}'21)}, pages 2280--2291. Curran Associates, 2021.

\bibitem[Moussa et~al.(2024)Moussa, Patel, Dunjko, B{\"a}ck, and van Rijn]{moussa-ml-24a}
C.~Moussa, Y.~J. Patel, V.~Dunjko, T.~B{\"a}ck, and J.~N. van Rijn.
\newblock Hyperparameter importance and optimization of quantum neural networks across small datasets.
\newblock \emph{Machine Learning}, 113\penalty0 (4):\penalty0 1941--1966, 2024.

\bibitem[Nauta et~al.(2023)Nauta, Trienes, Pathak, Nguyen, Peters, Schmitt, Schl{\"{o}}tterer, van Keulen, and Seifert]{nauta-cs23a}
M.~Nauta, J.~Trienes, S.~Pathak, E.~Nguyen, M.~Peters, Y.~Schmitt, J.~Schl{\"{o}}tterer, M.~van Keulen, and C.~Seifert.
\newblock From anecdotal evidence to quantitative evaluation methods: {A} systematic review on evaluating explainable {AI}.
\newblock \emph{{ACM} Comput. Surv.}, 55\penalty0 (13s):\penalty0 295:1--295:42, 2023.

\bibitem[Pedregosa et~al.(2011)Pedregosa, Varoquaux, Gramfort, Michel, Thirion, Grisel, Blondel, Prettenhofer, Weiss, Dubourg, Vanderplas, Passos, Cournapeau, Brucher, Perrot, and Duchesnay]{scikit-learn}
F.~Pedregosa, G.~Varoquaux, A.~Gramfort, V.~Michel, B.~Thirion, O.~Grisel, M.~Blondel, P.~Prettenhofer, R.~Weiss, V.~Dubourg, J.~Vanderplas, A.~Passos, D.~Cournapeau, M.~Brucher, M.~Perrot, and E.~Duchesnay.
\newblock Scikit-learn: Machine learning in {P}ython.
\newblock \emph{JMLR}, 12:\penalty0 2825--2830, 2011.

\bibitem[Probst et~al.(2019)Probst, Boulesteix, and Bischl]{probst-jmlr19a}
P.~Probst, A.~Boulesteix, and B.~Bischl.
\newblock Tunability: Importance of hyperparameters of machine learning algorithms.
\newblock \emph{JMLR}, 20\penalty0 (53):\penalty0 1--32, 2019.

\bibitem[Rasmussen and Williams(2006)]{rasmussen-book06a}
C.~Rasmussen and C.~Williams.
\newblock \emph{Gaussian Processes for Machine Learning}.
\newblock The MIT Press, 2006.

\bibitem[Rasulo et~al.(2024)Rasulo, Smith{-}Miles, Mu{\~{n}}oz, Handl, and L{\'{o}}pez{-}Ib{\'{a}}{\~{n}}ez]{rasulo-gecco24a}
A.~Rasulo, K.~Smith{-}Miles, M.~Mu{\~{n}}oz, J.~Handl, and M.~L{\'{o}}pez{-}Ib{\'{a}}{\~{n}}ez.
\newblock Extending instance space analysis to algorithm configuration spaces.
\newblock In \emph{Proceedings of the Genetic and Evolutionary Computation Conference Companion,}, pages 147--150. {ACM}, 2024.

\bibitem[Rodemann et~al.(2024)Rodemann, Croppi, Arens, Sale, Herbinger, Bischl, H{\"{u}}llermeier, Augustin, Walsh, and Casalicchio]{rodemann-arxiv24a}
J.~Rodemann, F.~Croppi, P.~Arens, Y.~Sale, J.~Herbinger, B.~Bischl, E.~H{\"{u}}llermeier, T.~Augustin, C.~Walsh, and G.~Casalicchio.
\newblock Explaining bayesian optimization by shapley values facilitates human-ai collaboration.
\newblock \emph{CoRR}, abs/2403.04629, 2024.

\bibitem[Sass et~al.(2022)Sass, Bergman, Biedenkapp, Hutter, and Lindauer]{sass-realml22a}
R.~Sass, E.~Bergman, A.~Biedenkapp, F.~Hutter, and M.~Lindauer.
\newblock Deepcave: An interactive analysis tool for automated machine learning.
\newblock In M.~Mutny, I.~Bogunovic, W.~Neiswanger, S.~Ermon, Y.~Yue, and A.~Krause, editors, \emph{{ICML} Adaptive Experimental Design and Active Learning in the Real World (ReALML Workshop 2022)}, 2022.

\bibitem[Segel et~al.(2023)Segel, Graf, Tornede, Bischl, and Lindauer]{segel-automl23a}
S.~Segel, H.~Graf, A.~Tornede, B.~Bischl, and M.~Lindauer.
\newblock Symbolic explanations for hyperparameter optimization.
\newblock In A.~Faust, C.~White, F.~Hutter, R.~Garnett, and J.~Gardner, editors, \emph{Proceedings of the Second International Conference on Automated Machine Learning}. Proceedings of Machine Learning Research, 2023.

\bibitem[Shahriari et~al.(2016)Shahriari, Swersky, Wang, Adams, and de~Freitas]{shahriari-ieee16a}
B.~Shahriari, K.~Swersky, Z.~Wang, R.~Adams, and N.~de~Freitas.
\newblock Taking the human out of the loop: {A} review of {B}ayesian optimization.
\newblock \emph{Proceedings of the {IEEE}}, 104\penalty0 (1):\penalty0 148--175, 2016.

\bibitem[Shapley(1953)]{shapley-book53a}
L.~S. Shapley.
\newblock \emph{17. A Value for n-Person Games}, pages 307--318.
\newblock Princeton University Press, Princeton, 1953.
\newblock ISBN 9781400881970.
\newblock \doi{doi:10.1515/9781400881970-018}.
\newblock URL \url{https://doi.org/10.1515/9781400881970-018}.

\bibitem[van Rijn and Hutter(2018)]{rijn-kdd18a}
J.~van Rijn and F.~Hutter.
\newblock Hyperparameter importance across datasets.
\newblock In Y.~Guo and F.~Farooq, editors, \emph{Proceedings of the 24th {ACM} {SIGKDD} International Conference on Knowledge Discovery and Data Mining ({KDD}'18)}, pages 2367--2376. ACM Press, 2018.

\bibitem[Wang and Perez(2017)]{wang-arxiv17a}
J.~Wang and L.~Perez.
\newblock The effectiveness of data augmentation in image classification using deep learning.
\newblock \emph{arXiv:1712.04621 [cs.CV]}, 2017.

\bibitem[Watanabe et~al.(2023)Watanabe, Bansal, and Hutter]{watanabe-ijcai-23a}
S.~Watanabe, A.~Bansal, and F.~Hutter.
\newblock Ped-anova: Efficiently quantifying hyperparameter importance in arbitrary subspaces.
\newblock In E.~Elkind, editor, \emph{Proceedings of the Thirty-Second International Joint Conference on Artificial Intelligence, {IJCAI-23}}, pages 4389--4396. International Joint Conferences on Artificial Intelligence Organization, 8 2023.
\newblock \doi{10.24963/ijcai.2023/488}.
\newblock URL \url{https://doi.org/10.24963/ijcai.2023/488}.
\newblock Main Track.

\end{thebibliography}

\end{document}